\documentclass[journal,draftclsnofoot,onecolumn,12pt]{IEEEtran}

\usepackage{graphicx}
\usepackage{amssymb}
\usepackage{booktabs}

\usepackage{lineno}
\usepackage{algorithm}
\usepackage{algorithmic}
\usepackage{subfigure}
\usepackage{multirow}
\usepackage{array}

\usepackage{epsfig}
\usepackage{graphicx}

\usepackage{adjustbox}
\usepackage{array}
\usepackage{booktabs}
\usepackage{colortbl}
\usepackage{float,wrapfig}
\usepackage{hhline}
\usepackage{multirow}

\usepackage{amsmath,amsfonts,amsthm,amssymb}
\usepackage{bm}
\usepackage{nicefrac}
\usepackage{microtype}

\usepackage{changepage}
\usepackage{extramarks}
\usepackage{fancyhdr}
\usepackage{lastpage}
\usepackage{setspace}
\usepackage{soul}
\usepackage{xspace}

\begin{document}



\title{I-WKNN: Fast-Speed and High-Accuracy WIFI Positioning for Intelligent Stadiums}





\author{Zhangzhi Zhao, Zhengying Lou, Ruibo Wang, Qingyao Li and Xing Xu
\thanks{Zhangzhi Zhao, Zhengying Lou, Ruibo Wang, Qingyao Li and Xing Xu are with University of Electronic Science and Technology of China, Chengdu, China} 
\vspace{-4mm}}
\maketitle

 

 

\begin{abstract}
Based on various existing wireless fingerprint location algorithms in intelligent sports venues, a high-precision and fast indoor location algorithm improved weighted k-nearest neighbor (I-WKNN) is proposed. In order to meet the complex environment of sports venues and the demand of high-speed sampling, this paper proposes an AP selection algorithm for offline and online stages. Based on the characteristics of the signal intensity distribution in intelligent venues, an asymmetric Gaussian filter algorithm is proposed. This paper introduces the application of the positioning algorithm in the intelligent stadium system, and completes the data acquisition and real-time positioning of the stadium. Compared with traditional WKNN and KNN algorithms, the I-WKNN algorithm has advantages in fingerprint positioning database processing, environmental noise adaptability, real-time positioning accuracy and positioning speed, etc. The experimental results show that the I-WKNN algorithm has obvious advantages in positioning accuracy and positioning time in a complex noise environment and has obvious application potential in a smart stadium.
\end{abstract}

\begin{IEEEkeywords}
Intelligent stadium, I-WKNN algorithm, WiFi fingerprint positioning, AP selection, asymmetric Gaussian filtering.
\end{IEEEkeywords}




\section{Introduction}
\label{S:1}

As early as the 2008 Beijing Olympic Games, the concept of Smart Sports Hall was put forward, which is the choice of sports modernization development and the choice of deepening scientific and technological innovation. Increasing the construction of intelligent stadiums can better meet the needs of sports competitions on the site. In 2014, China's State Council issued a No. 46 documents to encourage national fitness and vigorously build public sports infrastructure. With the continuous improvement of people's requirements for the quality of life, the traditional stadiums and gymnasiums only used for competition and sports can no longer meet people's needs. The emergence of intelligent stadiums and gymnasiums is an inevitable trend. An intelligent stadium system has a variety of functions, such as live events, sports data collection, and analysis, spectator navigation, and spectator statistics \cite{wang2016study}. Therefore, the intelligent stadium system based on position tracking can collect athletes' data faster, which significantly promotes sports information. It is of great practical significance to construct an intelligent gymnasium that can improve the scientific nature of physical education. First of all, the use of intelligent facilities can improve the scientific nature of sports. Secondly, sports venues are important places for all kinds of athletic activities. Compared with traditional sports venues, intelligent sports venues have inherent advantages in order management, risk protection, and other aspects, which can improve the order and safety of sports. Finally, intelligent technology can improve the use efficiency of sports venues, the utilization rate of physical education resources, reduce the waste of resources, and provide students with personalized physical exercise guidance and guide students to carry out sports activities with higher efficiency.

Location tracking is an essential problem in intelligent stadiums. At present, there are a lot of indoor positioning technologies based on different algorithms and technologies that have been proposed, such as ultra-wide bandwidth (UWB) positioning, Bluetooth positioning, WiFi positioning, etc. \cite{zhu2020indoor}. Ubisense proposed a UWB indoor positioning technology that is not affected by line-of-sight (LoS) and has high positioning accuracy \cite{s16050707}, but the hardware cost is very high. IBeacon is a protocol that can be used for indoor positioning systems proposed by Apple. It has low power consumption and comprehensive coverage, but it is easy to receive interference, leading to RSSI instability \cite{jianyong2014RSSI}. Microsoft proposed location fingerprint positioning technology based on WiFi. With the help of wireless local area network (WLAN) infrastructure, mobile intelligent terminal positioning can be realized without additional special equipment and only needs positioning software.

After years of construction, most large stadiums have completed the construction of a WiFi wireless network. WiFi technology provides an environmental basis for the application and development of intelligent sports construction. WiFi signal is undoubtedly a cheap and effective solution to complete positioning compared with other indoor positioning methods, such as sensors or Bluetooth, which often require additional equipment \cite{yang2012locating}. However, there are still many problems with intelligent stadium WiFi location technology. When serving athletes with fast movement speed and uncertain direction, receiving signal strength indicator (RSSI) of the access point (AP) with long-distance and unstable signal tends to fluctuate greatly \cite{jun2013social}. At the same time, in order to ensure positioning accuracy, the equipment often needs a faster sampling speed \cite{shu2015gradient} to ensure that there will be no noticeable calculation delay in the process of rapid data collection.

\subsection{Related work}
\label{S:2}
At present, indoor positioning technology based on WiFi is most suitable for large-scale use in intelligent stadiums. Wireless infrastructure has been widely deployed in all indoor places. The two key technologies of existing WiFi fingerprint locations are fingerprint database construction and fingerprint matching. The location design and collection of the initial fingerprint database usually adopt two modes. One is to locate reference points, and the corresponding fingerprint information through intensive collection of fingerprints in the location scene \cite{serodio2012lightweight}. Second, the location and fingerprint information of representative and relatively sparse reference points are collected. Then the fingerprint database is reconstructed by interpolating or predicting fingerprint information of other locations in the positioning scene through a particular model \cite{guo2019robust,nikitaki2013localization}. A method using user feedback and rating mechanism to update the fingerprint database is proposed, but it can not effectively identify valid information and error feedback \cite{luo2017smartphone}. Moreover, a crowd-sourcing fingerprint database updating method selectively extracts reliable information through a clustering algorithm to update fingerprint data, which is an effective method to improve the quality of the fingerprint database \cite{niu2015wicloc}.

For WiFi indoor positioning, the mainstream positioning method is the classical k-nearest neighbor (KNN), and its variants, such as weighted k-nearest neighbor (WKNN) \cite{miao2014novel,lin2014group}. The location effect of these algorithms depends on data acquisition. In addition, even for a very accurate database, these methods tend to rudely discard all data except the predicted location and its neighbors, leading to inaccurate positioning. Although these methods ensure low computation and real-time, the utilization efficiency of the database is very low, and the accuracy is also limited. 
\par
Since the nature of the localization problem still belongs to the classification problem, the neural network will become potential direct competitors \cite{abdullah2016k,chen2017indoor}. The intuitive idea is that a three-dimensional image can be constructed from the X-axis, Y-axis, and WiFi signal intensity within the physical plane. By processing the database in the way of image processing \cite{zhang2017discriminative} and taking the positioning position as the classification result, each classification output can make use of other data of the database, thus ensuring high precision \cite{zhang2017elastic-net}. The self-organizing map SOM neural network further eliminates the randomness of the mapping, provides the stability of the algorithm, and at the same time ensures the low delay of the algorithm \cite{zhang2018deterministic}. Although there is no application of SOM in indoor positioning, it will be a very potential solution if a sufficient amount of special database information in specific scenarios such as smart stadium is provided to enhance SOM's generalization ability. But for the trained neural network, the dimension of input and output is fixed, which determines the upper limit of the algorithm to a certain extent. Input imprecise RSSI will reduce accuracy, also difficult to reduce the amount of computation. The positioning accuracy entirely depends on the output dimension. For a large stadium, the system must ensure a very high output dimension and a very high delay or accept a very low accuracy.
\par
Unlike the previous two algorithms, I-WKNN combines the specific signal distribution model and proposes asymmetric Gaussian filtering to improve the generalization ability. As an algorithm that can meet the requirements of real-time and high precision, it obtains lower delay by discarding some useless information and improving the database's utilization rate to ensure higher accuracy.

\subsection{Contributions}
The main contributions of the paper can be summarized as follows:
\begin{itemize}
    \item This paper puts forward the application scheme of WiFi location in the sports field. As far as we know, this was the first time that indoor real-time positioning technology was applied to real-time positioning in intelligent stadiums.
    \item The scheme of AP signal selection is established. The scheme filters out the AP with an unstable signal value and corrects the relatively stable AP with a Gaussian filter.
    \item This paper proposes a new algorithm I-WKNN to improve the positioning accuracy and speed. This algorithm improves the speed and reliability of positioning, mainly by optimizing fingerprint collection methods and reducing the search space for fingerprint matching. 
    \item The algorithm's performance is analyzed and compared from the perspective of delay and positioning accuracy. The comparison algorithm is the triangulation algorithm, the conventional WKNN algorithm, which is based on a fingerprint database.
\end{itemize}

\section{System Model}
\subsection{WiFi fingerprint positioning principle}
The principle of WiFi fingerprint positioning technology is mainly to match the test point’s WiFi signal strength information with the geographic coordinates \cite{8255985}. The strength of the WiFi signal, described by RSSI, monotonically decreases with distance. There are three stages in WiFi indoor positioning, the preparation phase, the offline training stage, and the real-time positioning stage. Note that the AP signal selection mechanism is introduced in the latter two stages.

Preparation stage: Divide the stadium into blocks, a total of $M$ blocks, one block as a reference point and corresponding to a geographic coordinate. Moreover, $N$ WiFi devices are fixed indoors, and each WiFi device is an AP.

Offline training stage: Developers select reference points and collect multiple RSSI values from different APs at each reference point. Then record the geographic coordinates of test points, the MAC addresses of APs, and the RSSI values as fingerprints \cite{hu2018toward,hu2018experimental}. The RSSI value should be processed before real-time positioning \cite{bisio2014trainingless}. More specifically, $\overrightarrow{\text{RSSI}}_{m}\left( n \right)$  represents the RSSI value of the ${{m}^{{th}}}$ reference point corresponding to the ${{n}^{{th}}}$ AP and $\overrightarrow{r}_{m,n}$  represents the RSSI value at $S$ the time slot obtained by sampling the ${{n}^{{th}}}$ AP at the ${{m}^{{th}}}$ reference point. In the $S$ time slot, the WiFi signal strength RSSI values $\overrightarrow{r}_{m,n}$ of $N$ AP signals are received, and then AP signals are selected in this stage at each reference point. In the $S$ time slot, the WiFi signal strength RSSI values $\overrightarrow{r}_{m,n}$ of $N$ AP signals are received, and then AP signals are selected in this stage at each reference point. In the offline training stage, the AP signals with a high signal loss rate and strong signal jitter are eliminated first, and then the AP signals are corrected by asymmetric Gaussian filtering. After the selection of AP signals, the RSSI value $\overrightarrow{\text{RSSI}}_{m}\left( n \right)$  of $N$ APs corresponding to $M$ reference points in the stadium is obtained, and the value range of $N$ is $\left[ 1,N \right]$, and the value range of $M$ is $\left[ 1,M\right]$. The WiFi fingerprint of the ${{m}^{{th}}}$ reference point is stored in the fingerprint database together with $N$  RSSI value $\overrightarrow{\text{RSSI}}_{m}\left( n \right)$  
 and corresponding APs’ MAC addresses. 

Real-time positioning stage: Users measure the RSSI value of $N$ AP signals in real-time at the current position in the stadium and then select the AP signals in the real-time positioning stage. In this stage, the AP signals with a high signal loss rate and strong signal jitter are eliminated first, and then the AP signals are corrected by asymmetric Gaussian filtering. After that, the RSSI value of the AP signal selected online is matched with the WiFi fingerprint recorded in the database. Then, the location estimation algorithm is run according to the matching results to estimate the user's geographic coordinates. The specific process is shown in the following subsection.

\subsection{A typical fingerprint location algorithm: WKNN}

\begin{figure}[t]
    \centering
    \includegraphics[width=0.8\linewidth]{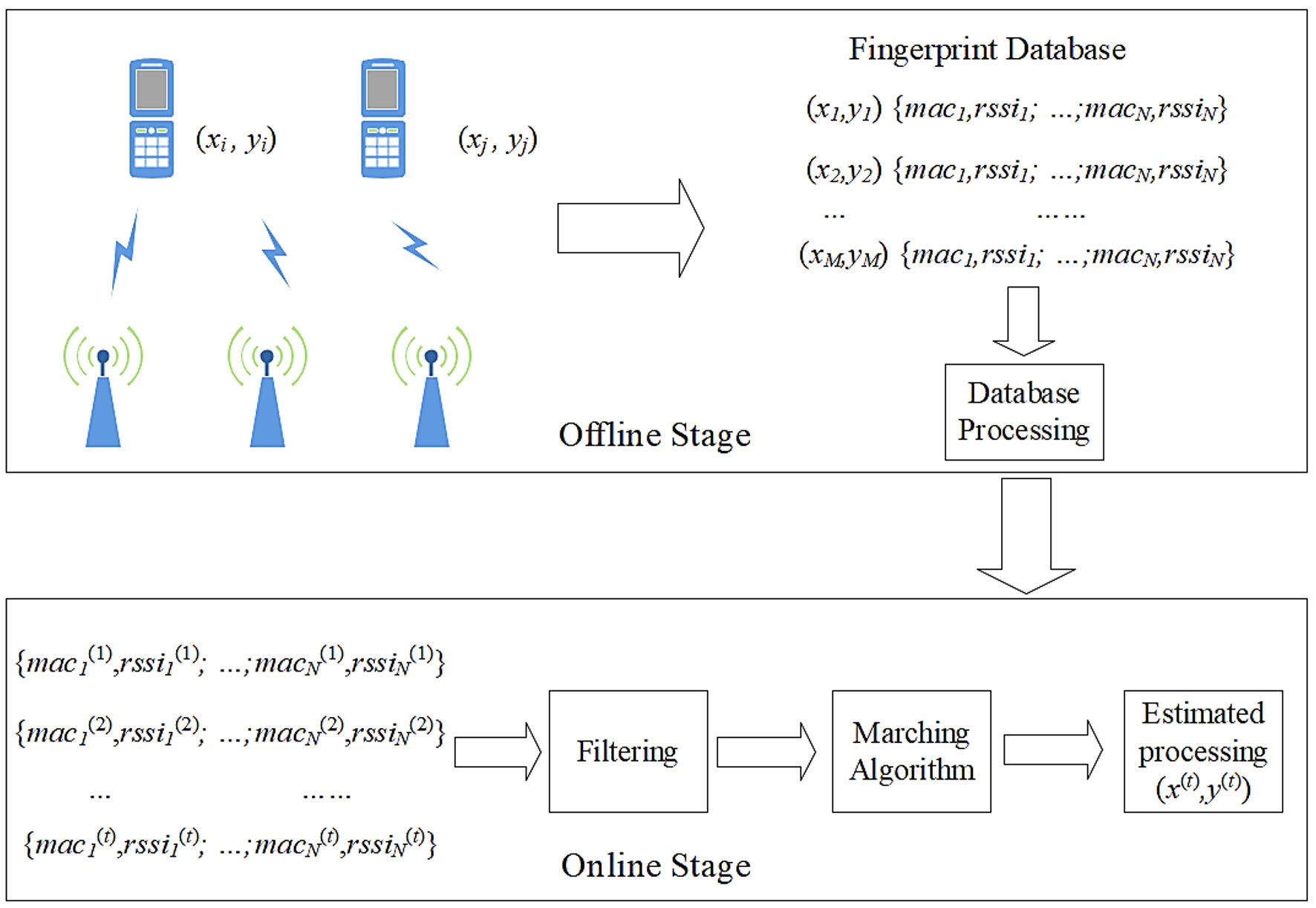}
    \caption{The Principle of WiFi Fingerprint Positioning.}
    \label{f1}
\end{figure}

The principle of WiFi fingerprint positioning technology matches the WiFi signal strength received by the test point with the geographic coordinates. The received WiFi signal strength is described by RSSI, which decreases monotonically with increasing distance. The WiFi indoor positioning is divided into two stages, and the specific process is shown in Figure~\ref{f1}.

Offline training stage: The developers divide the indoor area into blocks, and there are $M$ blocks in total. One block is a reference point and corresponding to a geographic coordinate $\left( {{x}_{i}},{{y}_{i}} \right),\text{ }i\text{ }=1,2,\cdots ,M$. Fixed $N$ WiFi devices indoors, and each WiFi device is an AP.  Multiple RSSI values for different APs are collected at each reference point. Then, the geographic coordinates of the reference point, the MAC address of APs, and the RSSI value are taken as the WiFi fingerprint of the reference point $\left( {{x}_{i}},{{y}_{i}} \right)$ is $\{{{\text{MAC}}_{1}},{{\text{RSSI}}_{1}};\ldots ;\text{ }{{\text{MAC}}_{N}},{{\text{RSSI}}_{N}}\}$, which is stored in the fingerprint database. 

Real-time positioning stage: The user (the object to be positioned) measure the RSSI  \\$\{{{\text{MAC}}_{1}}^{(1)},{{\text{RSSI}}_{1}}^{(1)};\ldots ;\text{ }{{\text{MAC}}_{N}}^{(1)},{{\text{RSSI}}_{N}}^{(1)}\}$, $\{{{\text{MAC}}_{1}}^{(2)},{{\text{RSSI}}_{1}}^{(2)};\ldots ; \text{ }{{\text{MAC}}_{N}}^{(2)},{{\text{RSSI}}_{N}}^{(2)}\}$, $\cdots $, $\{{{\text{MAC}}_{1}}^{(t)},{{\text{RSSI}}_{1}}^{(t)};\ldots ;\text{ }{{\text{MAC}}_{N}}^{(t)},{{\text{RSSI}}_{N}}^{(t)}\}$ at time slot $t$ in real-time at the current position. After filtering, the RSSI will carry out Matching Algorithm with the WiFi fingerprint recorded in the database. The location estimation algorithm is run to estimate the user's geographic coordinates based on the Matching results.

Then the WKNN algorithm is described in detail. At the time slot  $t$, analyze the geographical location of the stadium according to the positioning results of the last few moments. From the ${{i}^{{th}}}$ reference point in the fingerprint database, all of the physically adjacent reference points are selected, and the Euclidean geometric distance from them is calculated. The Euclidean geometric distance between the t moment and the ${{i}^{{th}}}$ reference point is denoted as $D(i,t)$,
\begin{equation}
D(i,t)=||{{\overrightarrow{\text{RSSI}}}_{i}}-\overrightarrow{\text{RSSI}}_{u}^{^{(t)}}|{{|}_{2}},
\end{equation}
where ${{\left\| \centerdot  \right\|}_{2}}$ represents the 2-norm of the vector, ${{\overrightarrow{\text{RSSI}}}_{i}}$ is the signal strength value of the ${{i}^{{th}}}$ reference point, which is an n-dimensional vector with an immutable order. $\overrightarrow{\text{RSSI}}_{u}^{^{(t)}}$ is the signal intensity value collected by the user at time slot $t$, and the vector is also n-dimensional, and the order should correspond to the order of APs in the database. If some APs’ RSSI values can not be measured, a small value $\text{RSSI}_{\min}$ is substituted (e.g. $-100$~dB). 

After obtaining the Euclidean geometric distance between the user and the ${{i}^{{th}}}$ reference point, we order the Euclidean distance values, leaving $K$ minimum values. Generally, the idea of weighted Euclidean algorithm in the Landmarc algorithm is adopted. The reciprocal of Euclidean distance square is used as the weighted factor, and the weight $w(k,t)$ of the reference point $i$ is denoted as,

\begin{equation}
w(k,t)=\left\{ \begin{matrix}
   \frac{1}{{{D}^{2}}(k,t)}/\sum\limits_{i}{\frac{1}{{{D}^{2}}(i,t)}}, & i\text{ is selected},  \\
   0, & \text{otherwise},  \\
\end{matrix} \right.
\end{equation}
the final positioning result is,
\begin{equation}
({{x}^{(t)}},{{y}^{(t)}})=\sum\limits_{i}{\text{w}(i,t) \cdot ({{x}_{i}},{{y}_{i}})},
\end{equation}
where $({{x}^{(t)}},{{y}^{(t)}})$ and $({{x}_{i}},{{y}_{i}})$ are all binary groups, representing the horizontal and vertical coordinate values in physical locations. However, in large venues, intelligent functions such as real-time positioning based on WKNN algorithm still face the problems of poor positioning accuracy and large positioning delay.

\subsection{Asymmetric Gaussian filtering}

For WiFi signal, if the distance between AP and user remains constant, the user receives a stable RSSI signal. However, when positioning in the stadium, the WiFi signal will be subject to any interference such as noise and blocking by athletes, and the intensity of the AP signal will also fluctuate. Assuming that the noise at each moment is independent of each other, and has a similar distribution, the stability of the RSSI time average can be approximately described by the Law of Large Numbers,
\begin{equation}
    \underset{n\to \infty }{\mathop{\lim }}\,\mathbb{P} \left[ \left| \frac{1}{n}\sum\limits_{i=1}^{n}{{\overrightarrow{\text{RSSI}}}_{\text{i}}}-\mu  \right|<\sigma  \right]=1,
\end{equation}
where $E({\overrightarrow{\text{RSSI}}})=\mu ,\text{ }D({\overrightarrow{\text{RSSI}}})={{\sigma }^{2}}\text{ }(i=1,2,\cdots )$.  Under the finite sample, it can be approximately considered that if the sample value is less than $\mu -3\sigma $ or greater than $\mu +3\sigma $, it can not be accepted. At the same time, the occurrence probability of this event is only around 0.3 percent.

RSSI is enhanced only because of the protocol in the measurement of the stadium signal, and it is not significant. However, because of complex scenes inside the stadium, depth decline often occurs, so it can be considered that the signal does not meet the Symmetry Gaussian distribution. The signal which is less than $\mu -{{g}_{\inf }}\sigma $ or greater than $\mu +{{g}_{\text{sup}}}\sigma $ should be eliminated, with g as the parameters to be determined. Therefore, the problem of Gaussian Filtering can be translated into,
\begin{equation}
    \left\{ \begin{matrix}
   \arg \min  & {{g}_{\text{inf}}}+{{g}_{\text{sup}}}  \\
   \text{s}.t. & \int_{\mu -{{g}_{\text{inf}}}\sigma }^{\mu +{{g}_{\text{sup}}}\sigma }{f(x)dx>1-\text{ }\!\!\epsilon\!\!\text{ }}  \\
\end{matrix} \right.,
\end{equation}
where $\epsilon$ is the allowable deviation of the algorithm, $f(.)$ is the probability density function of the sample.

In other words, the goal of asymmetric Gaussian filtering is to filter out as few signals as possible on the premise of tolerating errors. Since the problem is non-convex and requires extensive computation, this optimization problem is only used in the offline phase. In the online phase, we directly utilize the offline phase's solution results to reduce the solution time. 

\section{I-WKNN Algorithm}
\subsection{Offline stage of the Algorithm}

Assuming that there are S values of RSSI, they are obtained from sampling the ${{n}^{{th}}}$ AP at the $m$ reference point. They are denoted as ${{\overrightarrow{r}}_{m,n}}$. In the offline stage, if the signal is lost seriously or fluctuates violently, the data of the source will be removed.
When the signal cannot be measured, it will be replaced by a smaller value $\text{RSSI}_{\min}$. If the signal has not received data, the value will be zero after subtracting this value. The proportion of unreceived data can describe the degree of signal loss, that is to say, the ratio of 0-norm value to vector length in step (B). The elimination of the signal is replaced by $\text{RSSI}_{\min}$, which is equivalent to the data value when the signal is not received. The normalized variance in step (C) is used to describe the fluctuation. Asymmetric Gaussian filtering in step (D) eliminates partial signals that deviate entirely from the expected value. Then the AP selection algorithm in the offline stage is:
{
\begin{algorithm}[h]
    \caption{AP selection in offline phase}
    \begin{algorithmic}
    \STATE \textbf{(A) Initialize:} $m \leftarrow 1,n \leftarrow 1$, initialize ${{\Theta}_{1}}$ and ${{\Theta }_{2}}$.

    \FOR{m = 1 \textbf{to} M}
    \FOR{n = 1 \textbf{to} N}
    
    \STATE \textbf{(B) Eliminated part of AP by the signal loss rate:} 
    \IF{$\left \|{{\overrightarrow{\text{r}}}_{m,n}}+\text{RSSI}_{\min} \right \|_{0} \geq S{{\Theta }_{1}}$}
    \STATE ${{\overrightarrow{\text{RSSI}}}_{m}}(n) \leftarrow \text{RSSI}_{\min}$.
    \ENDIF
    
    \STATE \textbf{(C) Eliminated part of AP by fluctuation:}
    \IF{$\left \|\text{S}\cdot {{\overrightarrow{\text{r}}}_{m,n}} - \left \| {\overrightarrow{\text{r}}_{m,n}} \right \|_{1}\right \|_{2} \geq {{\Theta }_{2}} \left \|{{\overrightarrow{\text{r}}}_{m,n}}\right \|_{1}$}
    \STATE ${{\overrightarrow{\text{RSSI}}}_{m}}(n) \leftarrow \text{RSSI}_{\min}$.
    \ENDIF
    
    \STATE \textbf{(D) Perform asymmetric Gaussian filtering:}
    \IF{${{\overrightarrow{\text{RSSI}}}_{m}}(n)\ne \text{RSSI}_{\min}$}
    \STATE ccording to criterion of asymmetric Gaussian filtering, eliminate the RSSI value which is less than $\mu -{{g}_{\inf }}\sigma $ or larger than $\mu +{{g}_{\text{sup}}}\sigma $ in ${{\overrightarrow{\text{r}}}_{m,n}}$, then take the mean of the subsectors as ${{\overrightarrow{\text{RSSI}}}_{m}}(n)$.
    \ENDIF
    
    \ENDFOR
    \ENDFOR
    
\end{algorithmic}
\label{a1}
\end{algorithm}
\par
In step (A), ${{\Theta}_{1}}$ is the miss rate threshold,  and ${{\Theta }_{2}}$ is the jitter peak average ratio threshold. $\text{RSSI}_{\min}$ in step (B) is a minimum preset value. ${{\left\| \centerdot  \right\|}_{p}}$ is the p-norm of a vector, where $p=0,1,2$.
}

\subsection{I-WKNN Algorithm with Its Online stage}

In the offline stage, the parameter selection of the Gaussian filter is related to the data distribution in the database. To some extent, it requires human intervention. If the filtering effect is not effective, the parameters need to be changed. In the offline stage, each reference point will have a set of parameters of asymmetric Gaussian filtering for each AP. For the sake of saving time, this set of parameters will be directly used in the online stage. 

{
\begin{algorithm}[h]
    \caption{AP selection in offline phase}
    \begin{algorithmic}
    \STATE \textbf{(A) Initialize:} $n \leftarrow 1$, initialize the threshold ${{\Theta }_{1}}$ and ${{\Theta }_{2}}$.
    
    \FOR{n = 1 \textbf{to} N}
    \STATE \textbf{(B) Eliminated part of AP by the signal loss rate:} 
    
    \IF{$\left \|   \sum\limits_{\tau =t-T+1}^{t} {\overrightarrow{\text{RSSI}_{\text{u}}^{\left( \tau  \right)}}\left( n \right)}-\text{RSSI}_{\min}   \right \|_{0}\leq T{{\Theta }_{1}}$,$\tau \in [t-T+1,t]$}
    
        \STATE $\overrightarrow{\text{RSSI}_{\text{u}}^{\left( t \right)}}\left( n \right)=\frac{1}{T}\sum\limits_{\tau =t-T+1}^{t}{\overrightarrow{\text{RSSI}_{\text{u}}^{\left( \tau  \right)}}\left( n \right)}$,
        
        \ELSE
        \STATE Abandoned the $n^{th}$ AP information in time $\tau$.
        
    \ENDIF
    \ENDFOR
    
    \STATE \textbf{(D) obtain the RSSI of $N$ APs:} \\ $\overrightarrow{\text{RSSI}_{\text{u}}^{\left( t \right)}} \leftarrow \left( \overrightarrow{\text{RSSI}_{\text{u}}^{\left( t \right)}}\left( 1 \right),\overrightarrow{\text{RSSI}_{\text{u}}^{\left( t \right)}}\left( 2 \right),\cdots ,\overrightarrow{\text{RSSI}_{\text{u}}^{\left( t \right)}}\left( N \right) \right)$.
\end{algorithmic}
\label{a2}
\end{algorithm}
The algorithm assumes that the measurements at $T$ moments before the current moment are still valid. At time $T$, the AP selection algorithm in the online stage is as follows: In step (B), $T$ is the impact time on the measurement results at the time slot $t$, $\overrightarrow{\text{RSSI}}_{\text{u}}^{(\tau )}\left( n \right)$ is the signal strength received by the user at the time moment $\tau$, and the value range of $\tau$ is $[t-T+1,t]$. Step (B) filters out part of the signal values according to the signal strength values, filter the remaining signal according to the criterion of asymmetric Gaussian filtering. The ${{n}^{{th}}}$ AP information received by the user at time slot $t$ after asymmetric Gaussian filtering is the mean value of filtered results after $T$ time slots before $t$ time slot.

\par
In addition to standard WKNN, the algorithm introduces the AP selection mechanism and asymmetric Gaussian filtering algorithm. The overall flow chart of the I-WKNN algorithm is shown in Figure~\ref{f2}, containing the offline stage and online stage. 
\begin{figure}
    \centering
    \includegraphics[width=0.8\linewidth]{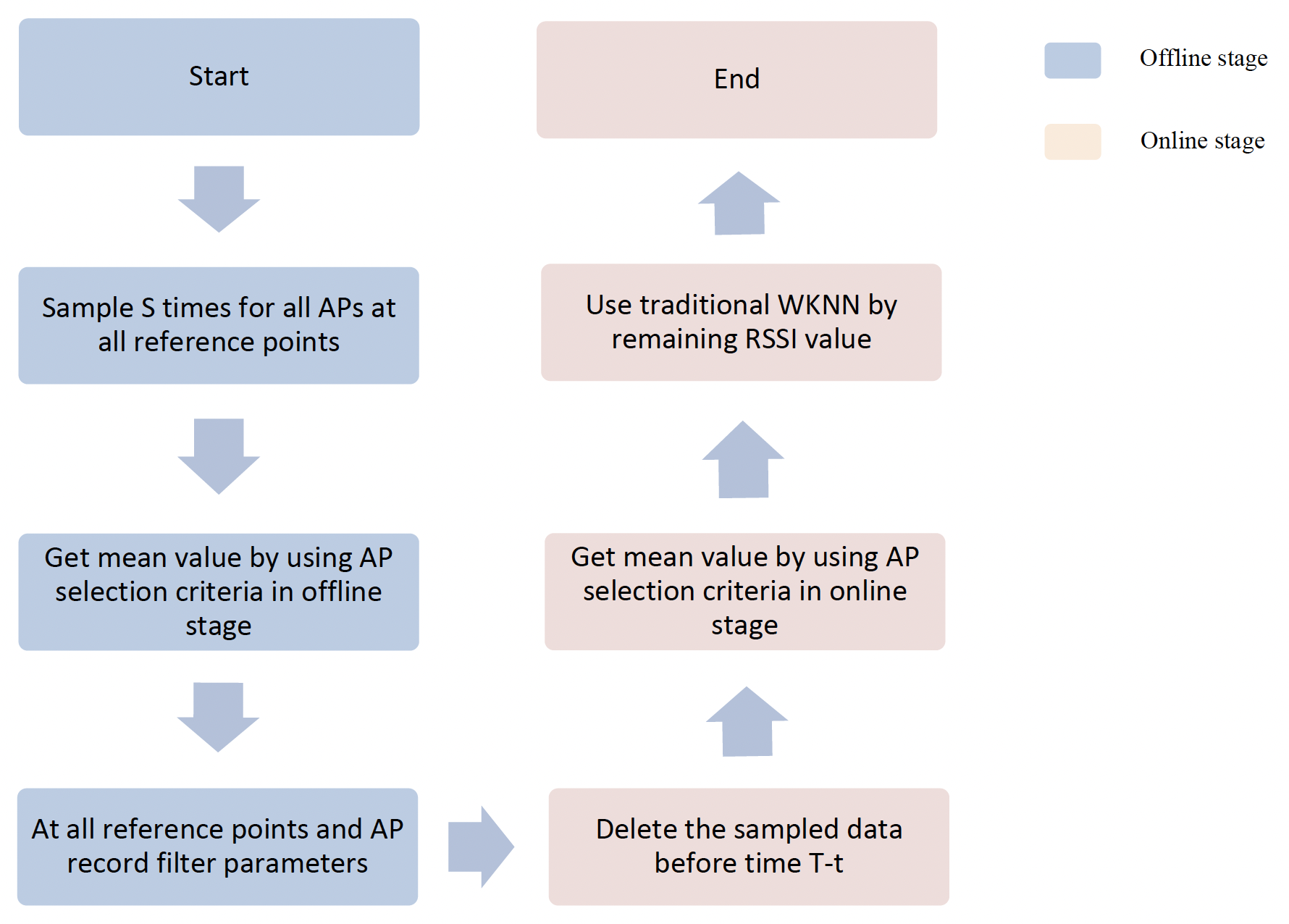}
    \caption{Flow Chart for Positioning.}
    \label{f2}
\end{figure}
}

\section{The Application in Intelligent Stadiums}

\subsection{Experimental Scene}

In this article, the intelligent stadium is based on the client/server (C/S) architecture, uses JAVA as the server development language, and MySQL database as the fingerprint database. Our server implements three main functions: it can efficiently complete the database storage and socket communication.

\begin{figure}
    \centering
    \includegraphics[width=0.8\textwidth]{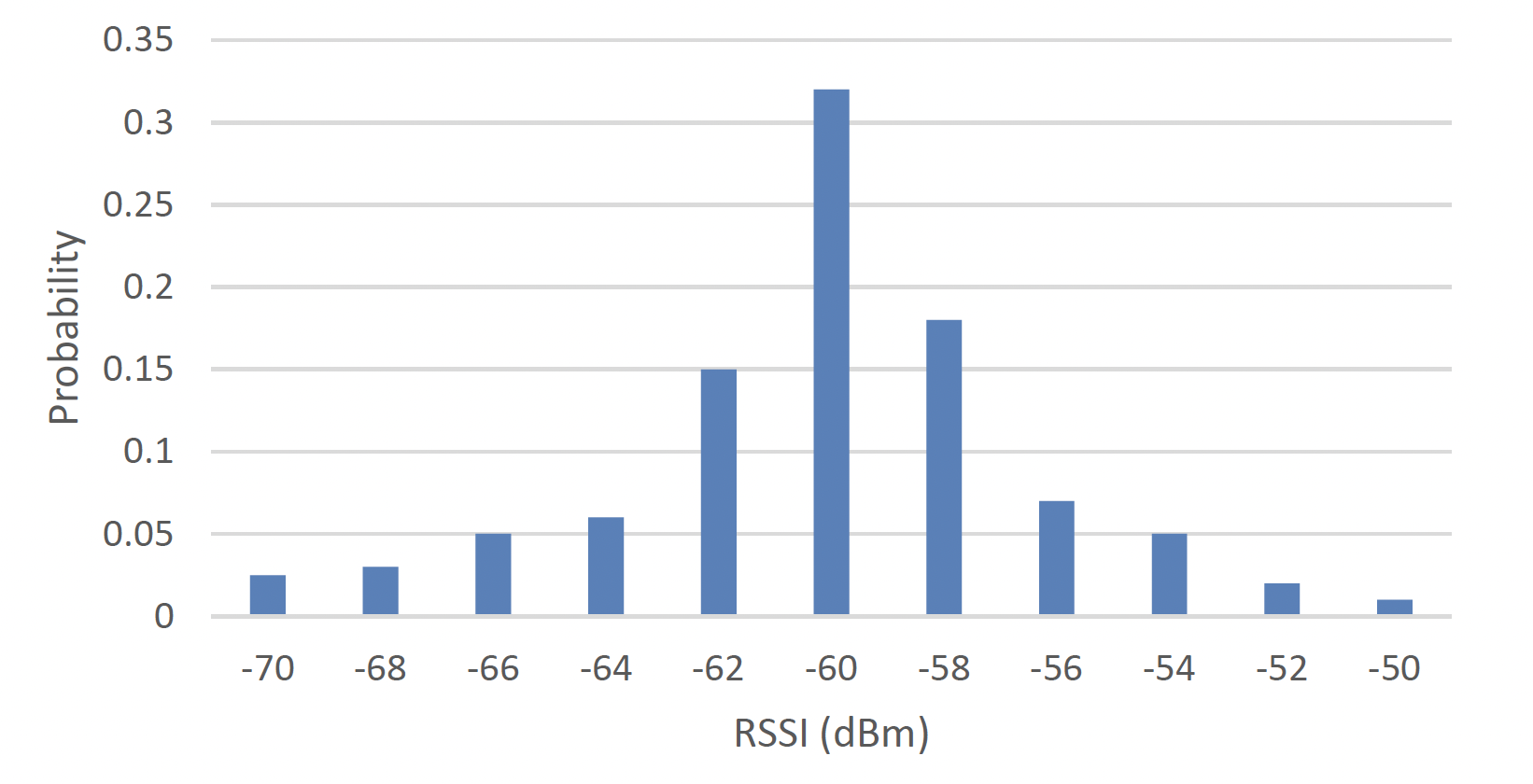}
    \caption{The Distribution of RSSI in An Ideal Situation.}
    \label{f3}
\end{figure}

The typical attenuation of WiFi signal is caused by the superposition of signal propagation in space. RSSI can present the effect of dynamic distribution, and the dynamic change and the nature of the most common Gaussian distribution are exceptionally similar under the free space. In addition, when pedestrians are excluded, the collection of RSSI presents a Gaussian distribution, as shown in Figure~\ref{f3}.

In an ideal channel, the data obtained by RSSI sampling at the same place at different times is approximately Gaussian. Under multipath channels, the distribution with maximum information entropy is Gaussian distribution. Under the condition of known mean value and variance, the Gaussian distribution model is first introduced. However, in the actual scene, the RSSI of the WiFi signal will fluctuate, superposition, and disappear due to the influence of shielding, personnel movement, and multipath in the sampling process. Thus, instead of presenting as a Gaussian distribution, it is the double-peak situation in the actual sampling.

Database and real-time measured RSSI information and geographical location are uploaded and processed by MATLAB software on the computer. We set up 250 points in the test area as clustering points, then numbered the position of each training point from right to left and from top to bottom. Run the WiFi location client program at each training point, measured each address and signal strength 1000 times. The data was processed, and after Gaussian filtering, the remaining records were written to the MySQL database.

\begin{figure}
    \centering
    \includegraphics[width=0.8\linewidth]{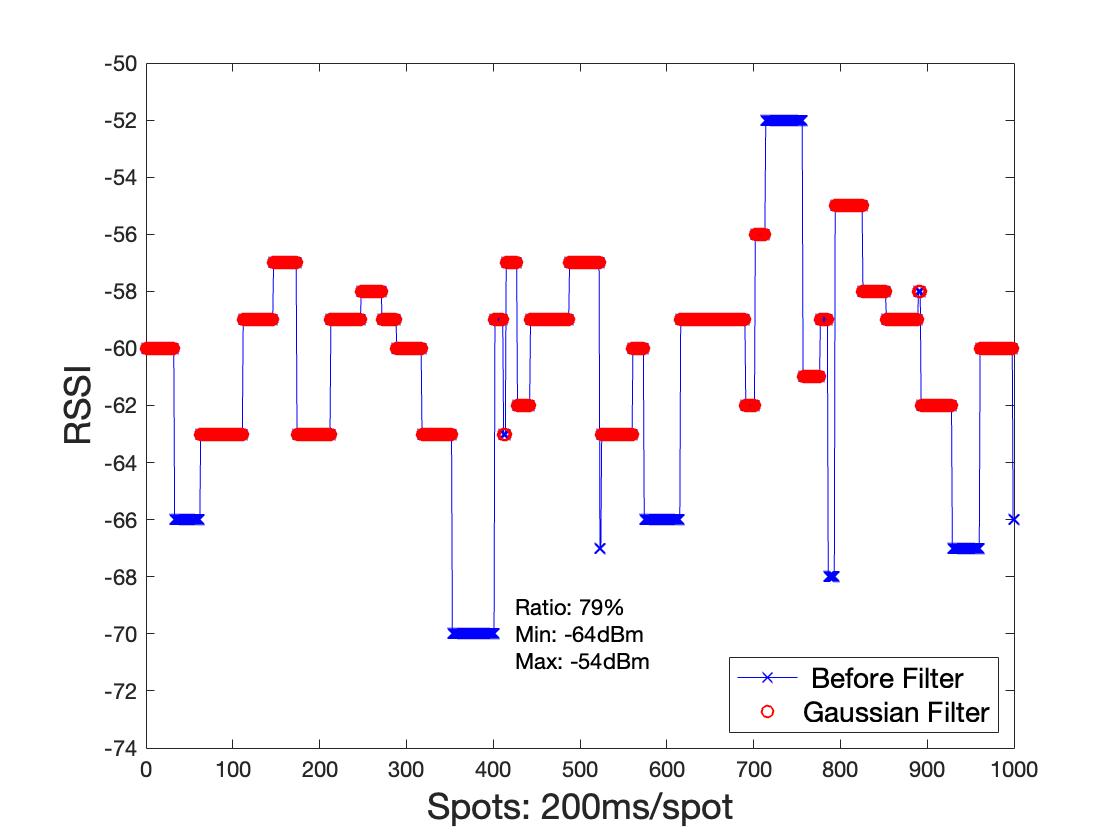}
    \caption{The Result of Gaussian Filtering for One AP.}
    \label{f5}
\end{figure}

The result of Gaussian filtering for one AP is shown in Figure~\ref{f5}. The blue data should be abandoned, and the red data will remain. From the signal distribution of the AP, the fading ratio is significantly higher than the enhancement ratio, and occasionally the signal cannot be detected, which is represented by $-70$~dB in this example.

\begin{figure}
    \centering
    \includegraphics[width=0.8\linewidth]{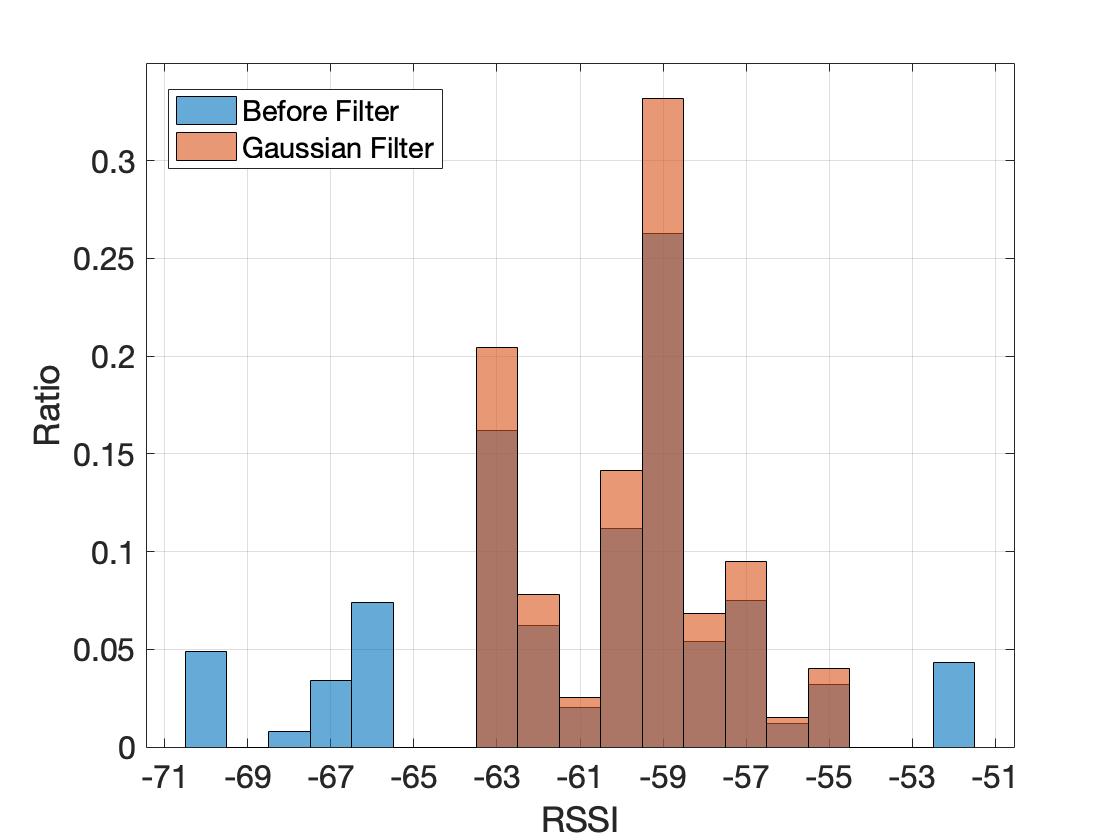}
    \caption{The Proportion of RSSI for One AP.}
    \label{f6}
\end{figure}

Moreover, we show the corresponding histogram of the proportion of RSSI in Figure~\ref{f6}. It can be seen that through the Gaussian filter, the signal becomes noticeably more concentrated. Moreover, the adaptive Gaussian filter can better describe the current RSSI distribution of WiFi signals.

\subsection{The Performance of Accuracy}

In the online stage, we set $K=5$ and placed 10 APs. The device stores information for the last 20 slots. The following two comparison algorithms are given.
\begin{itemize}
    \item WKNN algorithm: No extra processing for fingerprint database, just find five largest RSSI values, and get the gravity center of the corresponding five points in every moment.
    \item KNN algorithm: A sample belongs to a category if most of the K most similar (that is, closest to each other in the feature space) samples in the feature space belong to that category.
\end{itemize}

\begin{figure}
    \centering
    \includegraphics[width=0.8\linewidth]{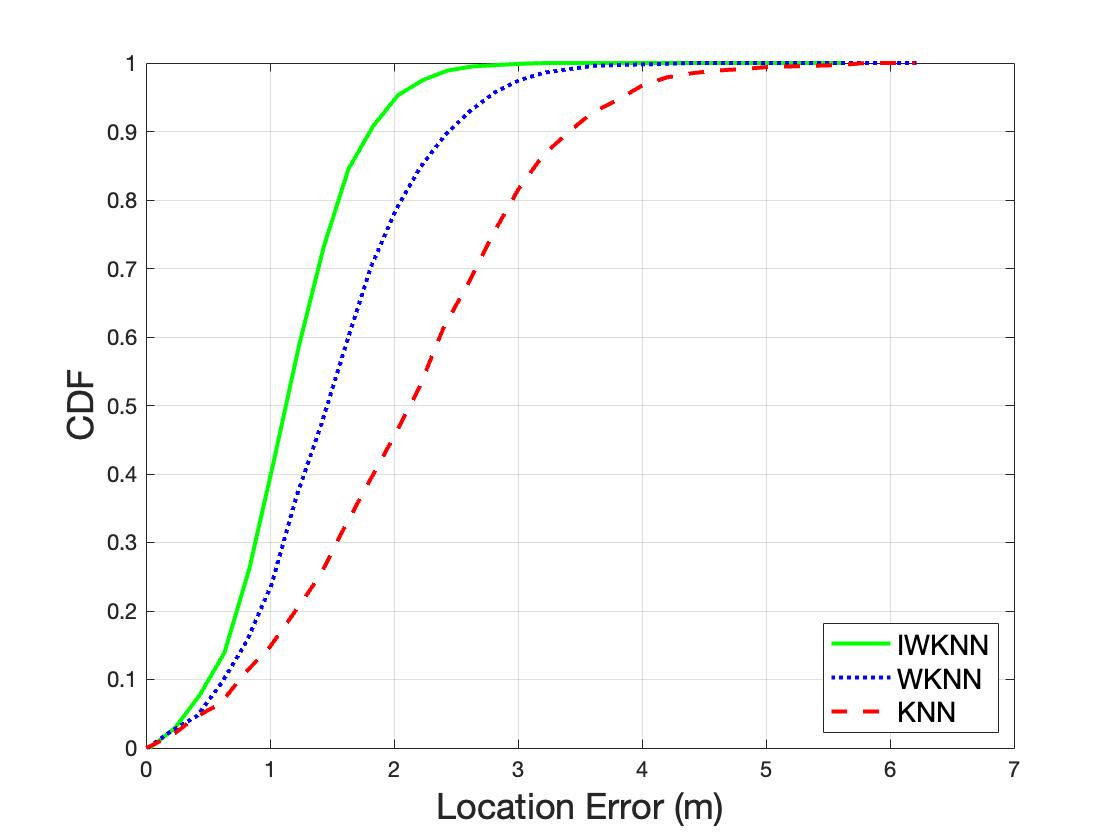}
    \caption{The CDF Graph of Accuracy for Three Algorithms.}
    \label{f7}
\end{figure}

The CDF graph is used to observe the accuracy of each positioning since the result of each positioning is equally important and the positioning error of each positioning needs to be known. It can be seen from the Figure~\ref{f7}, that the algorithm proposed in this paper is superior to the other two. Although the elimination of some APs seems to discard some information, the accuracy is greatly improved. In addition, it can be seen from the figure that the proportion of deviation of the four positioning algorithms below 2m is 95\% (I-WKNN), 79\% (WKNN), 39\% (WKNN), and 46\% (KNN), respectively. Similarly, the mean positioning deviations were 1.14~m (I-WKNN), 1.42~m (WKNN), and 2.32~m (KNN), respectively.

\begin{figure}
    \centering
    \includegraphics[width=0.8\linewidth]{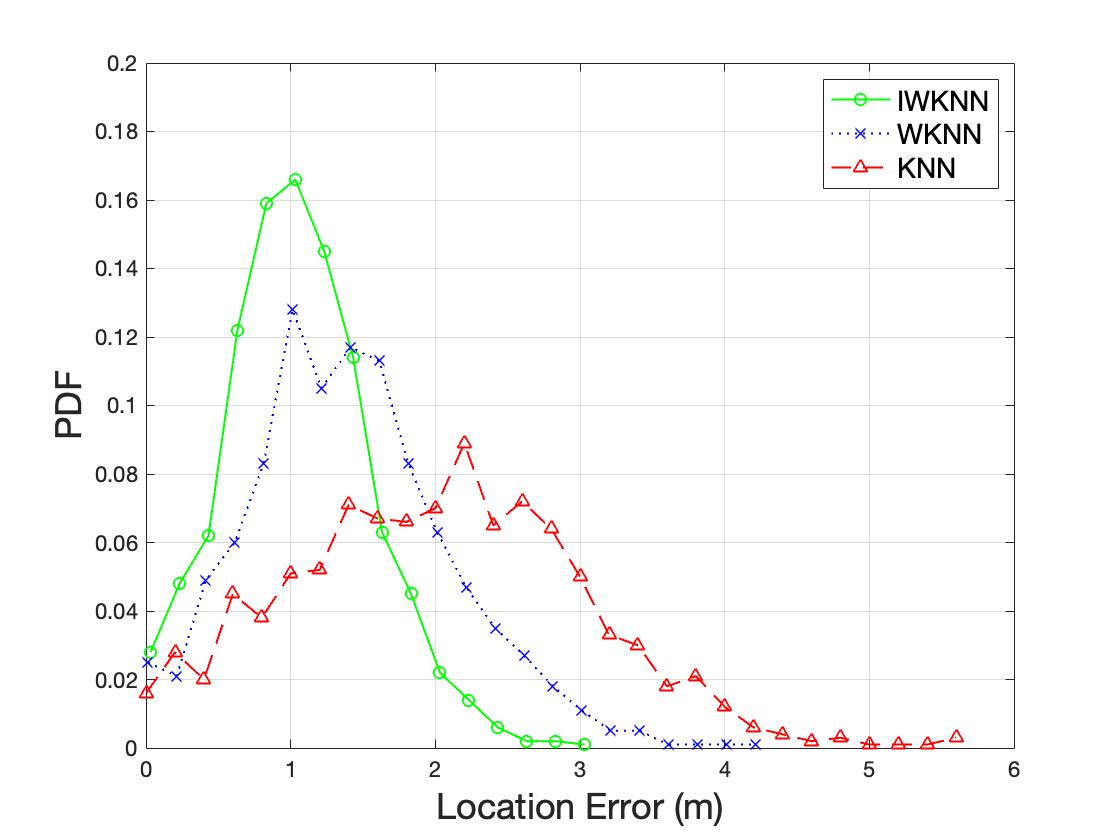}
    \caption{The PDF Graph of Accuracy for Three Algorithms.}
    \label{f8}
\end{figure}

The corresponding PDF of location error is shown in Figure~\ref{f8}. The proposed method has the highest probability of location error of about 1~m, and the maximum error will not exceed 3~m. The error of WKNN is slightly larger than that of our method. The performance of KNN is the worst, with a maximum error of almost 6~m.

\subsection{The Performance of Time Delay}

\begin{figure}
    \centering
    \includegraphics[width=0.8\linewidth]{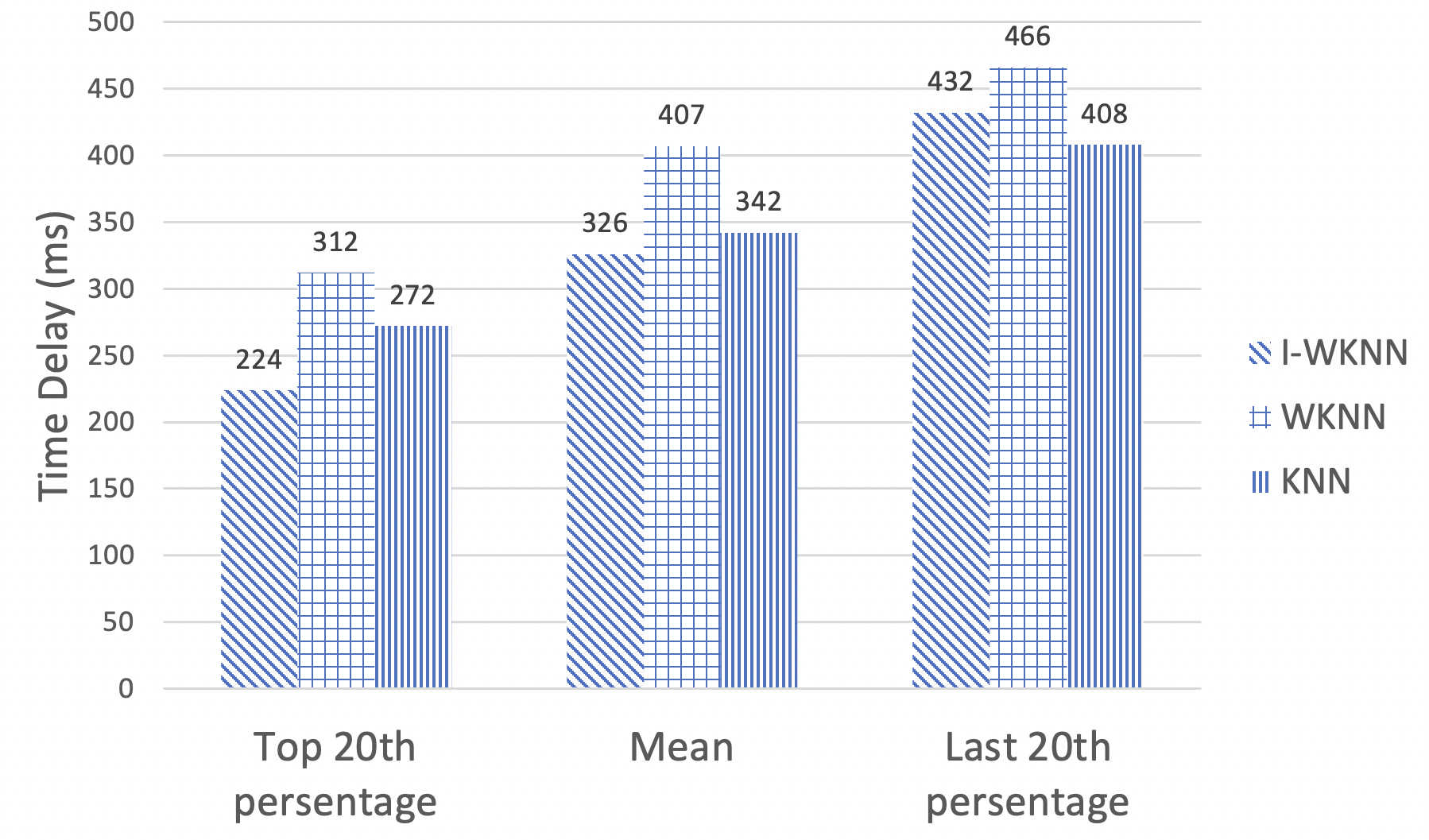}
    \caption{The Comparison of Average Time Delay for Three Algorithms.}
    \label{f9}
\end{figure}

To reduce the influence of various factors such as device heterogeneity and time uncertainty on the positioning accuracy, we used a single device to run the two algorithms and completed 40 positioning. The overall time-consuming and average time-consuming of the experiment are shown in Figure~\ref{f9}. Because positioning speed is related to hardware and software performance, this chart is for reference only.

Unlike the positioning accuracy, the time delay only needs to be lower than an acceptable value, and the samples with a significant delay should be paid special attention. So the bar chart shows the average delay, the best 20\% experiment, and the worst 20\% experiment, rather than every sample.

As can be seen from the figure, the KNN method, which has a significant delay, is entirely unsuitable for the rapidly changing scene such as an intelligent stadium. For the best cases, the delay of the I-WKNN algorithm has the smallest time delay. In terms of average delay, I-WKNN has obvious improvement over traditional WKNN. For the worst 20 percent, I-WKNN is still a small advantage over WKNN.

\section{Conclusion}

An improved WKNN algorithm is proposed in this paper, called I-WKNN. The improved AP selection algorithm and asymmetric Gaussian filter algorithm optimize the offline and online stage of fingerprint location. In the experiment, the triangulation algorithm, the traditional WKNN algorithm, is used to compare the proposed one. The accuracy of I-WKNN has obvious advantages compared with the other three algorithms. Its average deviation is 1.14~m, and the proportion of the deviation lower than 2~m is 95\%. I-WKNN algorithm is only worse than triangulation in the time delay, thus ranks second, with an average delay of 326~ms and a maximum delay of 432~ms, which can meet the requirements of rapid positioning of the stadium. It can be seen that in complex environments, it has high precision and fast positioning speed, which will make it suitable for the scene of the stadium.



\bibliographystyle{IEEEtran}
\bibliography{ccs}

\end{document}